\def\year{2022}\relax
\documentclass[letterpaper]{article} 
\usepackage{arxiv}  
\usepackage{times}  
\usepackage{helvet}  
\usepackage{courier}  
\usepackage[hyphens]{url}  
\usepackage{graphicx} 
\usepackage{natbib}  
\usepackage{caption} 
\DeclareCaptionStyle{ruled}{labelfont=normalfont,labelsep=colon,strut=off} 
\usepackage{algorithm}
\usepackage{algorithmic}

\usepackage{newfloat}
\usepackage{listings}
\floatstyle{ruled}
\newfloat{listing}{tb}{lst}{}
\floatname{listing}{Listing}

\usepackage{bm}
\usepackage{physics}
\usepackage{amsfonts}
\DeclareMathOperator{\sign}{sign}
\DeclareMathOperator*{\argmax}{arg\,max}

\title{Adversarially Trained Object Detector for Unsupervised Domain Adaptation}

\author {
    Kazuma Fujii,\textsuperscript{\rm 1}
    Hiroshi Kera, \textsuperscript{\rm 2}
    Kazuhiko Kawamoto \textsuperscript{\rm 2}
}
\affiliations {
    \textsuperscript{\rm 1} Graduate School of Science and Engineering, Chiba University\\
    \textsuperscript{\rm 2} Graduate School of Engineering, Chiba University\\
    \{fujisan8, kera\}@chiba-u.jp, kawa@faculty.chiba-u.jp
}

\begin{document}

\maketitle

\begin{abstract}
Unsupervised domain adaptation, which involves transferring knowledge from a label-rich source domain to an unlabeled target domain, can be used to substantially reduce annotation costs in the field of object detection. 
In this study, we demonstrate that adversarial training in the source domain can be employed as a new approach for unsupervised domain adaptation. 
Specifically, we establish that adversarially trained detectors achieve improved detection performance in target domains that are significantly shifted from source domains. 
This phenomenon is attributed to the fact that adversarially trained detectors can be used to extract robust features that are in alignment with human perception and worth transferring across domains while discarding domain-specific non-robust features. 
In addition, we propose a method that combines adversarial training and feature alignment to ensure the improved alignment of robust features with the target domain. 
We conduct experiments on four benchmark datasets and confirm the effectiveness of our proposed approach on large domain shifts from real to artistic images. 
Compared to the baseline models, the adversarially trained detectors improve the mean average precision by up to 7.7\%, and further by up to 11.8\% when feature alignments are incorporated.
Although our method degrades performance for small domain shifts, quantification of the domain shift based on the Fr\'echet distance allows us to determine whether adversarial training should be conducted.
\end{abstract}

\section{Introduction}
\label{sec:introduction}
In the field of computer vision, object detection is a fundamental task, which involves localizing and classifying objects in an image. 
Advancements in deep learning have resulted in various types of object detectors being proposed~\cite{girshick2015fast,ren2015faster,liu2016ssd,redmon2018yolov3,zhao2019m2det,tan2020efficientdet}. 
In most cases, they require supervised learning on a large amount of annotated data~\cite{everingham2010pascal,lin2014microsoft}.
Furthermore, to achieve the expected performance, the training and test data must belong to the same domain.

However, domain shifts resulting from changes in weather, painting style, and other factors often occur in practical applications, thereby resulting in a loss of accuracy.
In object detection tasks, during annotation, bounding boxes are required for all the objects in the images. 
Therefore, creating a new training dataset in the shifted domain is impractical.
An effective solution to this issue is domain adaptation, which involves transferring knowledge from a label-rich source domain to a label-poor or unlabeled target domain~\cite{ganin2016domain-adversarial}.
Specifically, unsupervised domain adaptation assumes that the target domain has no labels~\cite{zhao2020review}.
In recent studies, several approaches have been proposed for implementing unsupervised domain adaptation in object detection tasks~\cite{oza2021unsupervised}.
The most common approach is adversarial feature learning, which involves aligning the source and target features using a feature extractor competing with a domain discriminator~\cite{chen2018domain,saito2019strong-weak,chen2021scale}. 
Other approaches, such as pseudo-labeling in the target domain~\cite{kim2019self-training,li2021free} and image-to-image translation~\cite{hsu2020progressive,chen2020harmonizing} have also been proposed.

\begin{figure}[t]
\centering
\includegraphics[width=0.48\textwidth]{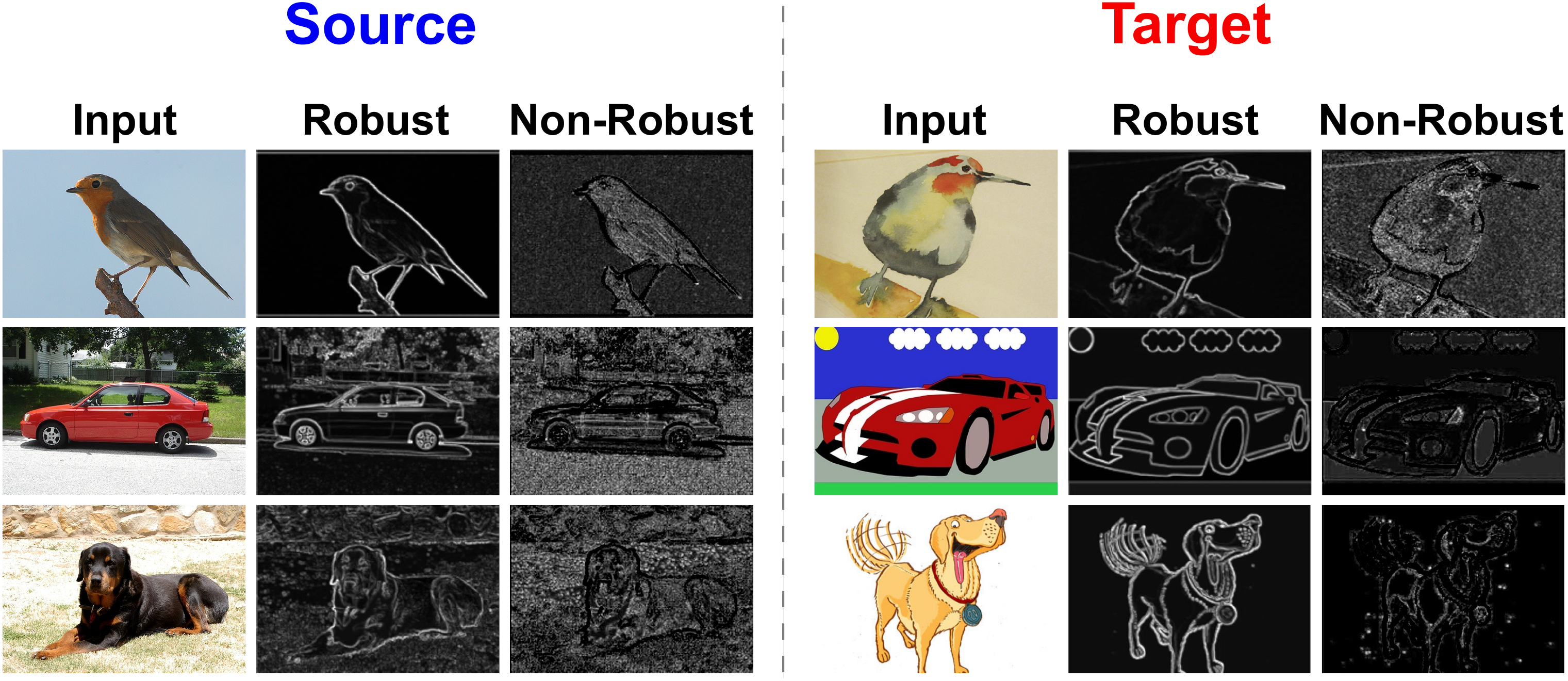}
\caption{
Visualization of robust and non-robust features.
The standard trained model in the source domain is highly dependent on non-robust features, which are not informative for the largely shifted target domain. In contrast, the robust features acquired by the adversarially trained model are informative even in the largely shifted target domain. 
}
\label{fig:fujii1}
\end{figure}

In this study, we explore the application of unsupervised domain adaptation in the field of object detection. Further, we demonstrate that the learning of robust features in the source domain through adversarial training enhances object detection in the target domain with a large domain shift.
Recent studies on adversarial training have revealed the existence of non-robust and robust features~\cite{ilyas2019adversarial}. 
The former are sensitive to perturbation, but they are still necessary for attaining high accuracy. 
The latter are highly stable and close to human perception~\cite{tsipras2019robustness}. 
Robust and non-robust features are visualized in Fig.~\ref{fig:fujii1}.
We hypothesize that for domain adaptation, non-robust features are highly domain-specific features, and thus, they are susceptible to domain shifts, whereas robust features are informative in both the source and target domains.
This idea is inspired by studies that have recently shown that adversarially trained models demonstrate improved transfer performance compared to standard-trained models~\cite{salman2020do,utrera2021adversarially}.
These studies focus on transfer learning in cases where the target domain has a small number of labels. 
Contrarily, we focus on unsupervised domain adaptation, where the target domain has no labels. 
In addition to learning robust features through adversarial training, to ensure the increased alignment of such features with the target domain, we propose a novel approach that combines adversarial training and adversarial feature learning~\cite{saito2019strong-weak}.

In our experiments on the benchmark datasets of real to artistic image adaptation, the adversarially trained detector improves the mean average precision by up to 7.7\% compared to that of the standard-trained detector.
When combined with adversarial feature learning, the improvement in mean average precision reaches 11.8\%.
Although our method degrades the performance for small domain shifts such as different weather conditions, quantifying the domain shift using the Fr\'echet distance allows us to predict domain adaptation performance with adversarial training in advance.
In addition, we analyze various adversarial training methods for object detection.
We demonstrate that several proposed techniques that have been suggested to be robust against adversarial examples are not substantially different from the simplest adversarial training method in terms of their application in unsupervised domain adaptation.

The contributions of this study are as follows:
\begin{itemize}
\item To the best of our knowledge, this is the first study on the effectiveness of adversarial training in unsupervised domain adaptation. 
We establish that, for large domain shifts, adversarially trained detectors achieve improved accuracy in the target domain compared to standard-trained detectors.
\item We propose a method that combines adversarial training with adversarial feature learning to ensure the enhanced alignment of the source and target features. 
Experimental results show that our proposed method achieves improved domain adaptation performance compared to approaches that solely rely on adversarial training.
\item We introduce a quantification of the domain shift using the Fr\'echet distance, which allows us to predict the domain adaptation performance with adversarial training.
\item We show that several adversarial training methods that have been proposed to improve robustness against adversarial examples do not differ substantially in terms of performance with respect to unsupervised domain adaptation.

\end{itemize}

\section{Related Work}
In this section, we review the literature pertaining to studies on object detection, domain adaptation, and adversarial training.
\subsection{Object detection}
Object detection is a fundamental task in computer vision as well as image classification.
Many object detectors have achieved high accuracy due to advancements in deep neural networks~\cite{girshick2015fast,ren2015faster,liu2016ssd,redmon2018yolov3,zhao2019m2det,tan2020efficientdet}.
Most of them rely on supervised learning using large annotated datasets, such as PASCAL Visual Object Classes (VOC)~\cite{everingham2010pascal} and Microsoft Common Objects in Context
(MSCOCO)~\cite{lin2014microsoft}.
Generally, creating a new dataset for object detection is more time-consuming than creating one for image classification because it requires instance-level annotations.
In this study, we use You Only Look Once v3 (YOLOv3), which is a well-known object detector with excellent inference speed and accuracy~\cite{redmon2018yolov3}.

\subsection{Domain adaptation}
Domain adaptation is a technique for adapting a model trained using a label-rich domain to a label-poor domain.
Recently, unsupervised domain adaptation has attracted significant attention in computer vision tasks, such as image classification and semantic segmentation~\cite{zhao2020review}.

Many domain adaptation approaches have also been proposed for object detection~\cite{oza2021unsupervised}.
Typical approaches include adversarial feature learning~\cite{chen2018domain,saito2019strong-weak,chen2021scale}, pseudo-label-based self-training~\cite{kim2019self-training,li2021free}, and image-to-image translation~\cite{hsu2020progressive,chen2020harmonizing}.
Adversarial feature learning employs an adversarial objective between the domain discriminator and feature extractor~\cite{ganin2016domain-adversarial}.
The domain discriminator attempts to accurately classify the source and target images, whereas the feature extractor attempts to fool the domain discriminator.
As a result, the model can extract similar features from the source and target domains. 
The pseudo-label-based self-training approach trains the model by assigning pseudo-labels to the target images based on the knowledge obtained from the source domain.
Image-to-image translation converts the source images into target-like images using CycleGAN~\cite{zhu2017unpaired} or similar methods.
The model is then trained using the converted images and the original labels obtained from the source domain.

We propose a new method based on adversarial training for unsupervised domain adaptation in object detection.
Moreover, our proposed method can be combined with adversarial feature learning, which is the most common approach.

\subsection{Adversarial training}
One of the vulnerabilities of deep neural network-based models is the existence of adversarial examples that perturb the inputs and cause such models to make mistakes~\cite{szegedy2014intriguing}.
During adversarial training, a model is trained using adversarial examples generated from the training data to ensure that the model is robust against input perturbations.
Note that adversarial training is different from adversarial feature learning.
The most typical methods for creating adversarial examples are the fast gradient sign method (FGSM)~\cite{goodfellow2015explaining} and the projected gradient descent (PGD)~\cite{madry2018towards}.
Such methods mainly focus on image classifiers. 
However, studies on the adversarial training of object detectors have also been conducted from the perspective of multi-task learning in object detection~\cite{zhang2019towards,chen2021robust}.

In a recent study, the researchers demonstrated that adversarial examples result from the presence of non-robust features that are highly predictive but imperceptible to humans~\cite{ilyas2019adversarial}.
Standard-trained models rely on such non-robust features, whereas adversarially trained models extract robust features that are aligned with human perception~\cite{tsipras2019robustness}.
This attribute gave rise to an unintended but useful inference, i.e., adversarially trained models are highly effective in transferring knowledge to new domains compared to standard-trained models~\cite{salman2020do,utrera2021adversarially}.

Inspired by the observations presented above, we propose adversarial training in the source domain as an approach for implementing unsupervised domain adaptation in object detection.
The robust features acquired from the source domain are informative in the dissimilar target domain.
Furthermore, the enhanced alignment of robust features with the target domain can be achieved through adversarial training combined with adversarial feature learning.

\section{Proposed Method}
\begin{figure}[t]
\centering
\includegraphics[width=0.45\textwidth]{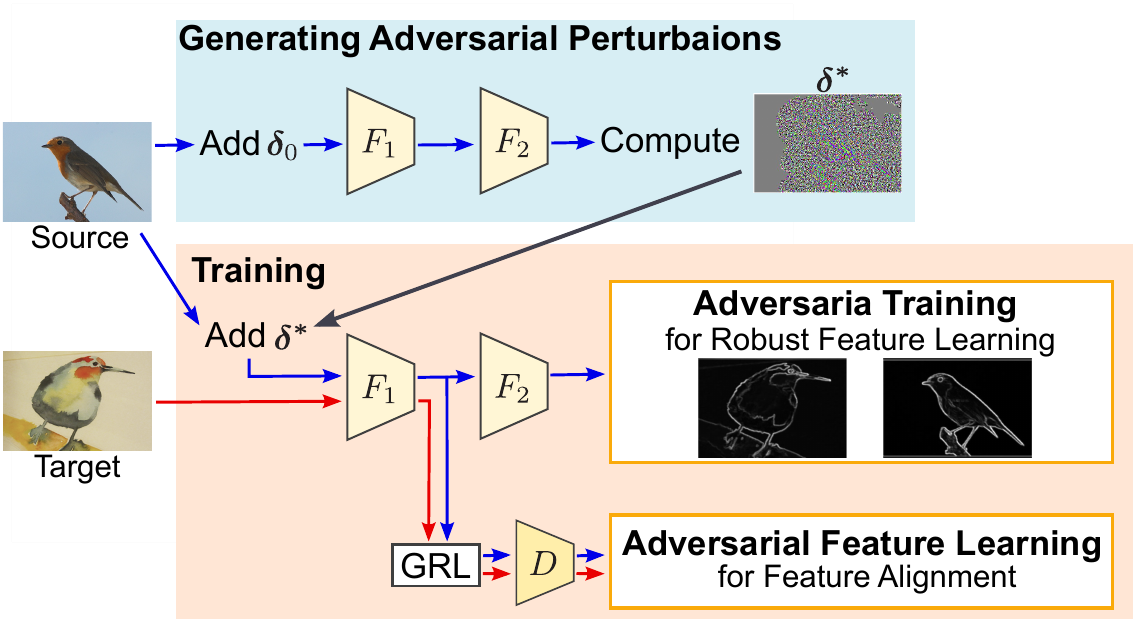}
\caption{
Framework of the proposed method.
$F_1$ and $F_2$ are object detection networks, $D$ is a domain discriminator, and GRL is a gradient reversal layer.
First, we propagate the source images with the initial perturbations $\bm{\delta}_{0}$ and then compute the adversarial perturbations $\bm{\delta^*}$ using the gradients of the losses.
Then, adversarial training on the source images perturbed by $\bm{\delta^*}$ and adversarial feature learning on the source and target images are performed.
}
\label{fig:fujii2}
\end{figure}
In this section, we first formulate the problem and describe the adversarial training in the source domain for YOLOv3~\cite{redmon2018yolov3}. 
We then introduce an approach for combining adversarial training and adversarial feature learning to ensure robust and target-aligned feature acquisition.
The framework of our proposed method is illustrated in Fig.~\ref{fig:fujii2}.

\subsection{Problem setting}
To implement unsupervised domain adaptation in object detection, we obtain labeled data $\qty(\bm{x}_s, \qty{y_s, \bm{b}_s})$ from the source domain $\mathcal{D}_s$ and unlabeled data $\bm{x}_t$ from the target domain $\mathcal{D}_t$.
Here, $\bm{x}_s$ and $\bm{x}_t$ represent the input images, $y_s$ represents the class label, and $\bm{b}_s$ represents the bounding box.
Generally, two domains, $\mathcal{D}_s$ and $\mathcal{D}_t$, have different data distributions.
The goal of domain adaptation is to improve the detection performance in the target domain $\mathcal{D}_t$ using the labeled data in the source domain and the unlabeled data in the target domain.
To avoid notational clutter, we use $\mathcal{D}_s$ and $\mathcal{D}_t$ to denote the data distributions of the source and target domains, respectively.

In this study, we use YOLOv3, which is a well-known object detector.
The objective of standard training in the source domain for YOLOv3 can be expressed as follows:
\begin{align}
    \min_{F} \mathbb{E}_{\qty(\bm{x}_s, \qty{y_s, \bm{b}_s})\sim\mathcal{D}_s}
    \qty[\mathcal{L}_{\mathrm{det}}\qty(F\qty(\bm{x}_s), \qty{y_s, \bm{b}_s})],
\end{align}
where $\mathcal{L}_{\mathrm{det}}$ denotes the detection loss, and $F$ denotes the YOLOv3 network.
Because $F$ outputs the class prediction, bounding box prediction, and objectness score, $\mathcal{L}_{\mathrm{det}}$ can be decomposed into the classification loss, localization loss, and objectness loss as follows:
\begin{align}
\label{eq:detection_loss}
  &\mathcal{L}_{\mathrm{det}}\qty(F\qty(\bm{x}_s), \qty{y_s, \bm{b}_s})  \notag\\ 
  &= 
  \mathcal{L}_{\mathrm{cls}}\qty(F\qty(\bm{x}_s), y_s) 
  + \mathcal{L}_{\mathrm{loc}}\qty(F\qty(\bm{x}_s), \bm{b}_s) 
  + \mathcal{L}_{\mathrm{obj}}\qty(F\qty(\bm{x}_s)).
\end{align}
Here, $\mathcal{L}_{\mathrm{cls}}$ is used to measure the difference between the predicted and ground-truth classes, $\mathcal{L}_{\mathrm{loc}}$ is used to measure the misalignment between the predicted and ground-truth boxes, and $\mathcal{L}_{\mathrm{obj}}$ is used to verify the existence of the predicted objects.

\subsection{Adversarial training in the source domain}
\label{subsec:at_in_source}
Our main objective is to demonstrate that adversarial training in the source domain can be employed as an approach for achieving unsupervised domain adaptation.
The robust features acquired through adversarially trained detectors are expected to be useful for dissimilar target domains and improve detection accuracy in the target domain.
The objective of adversarial training can be expressed as follows:
\begin{align}
    \label{eq:at}
    \min_{F} \mathbb{E}_{\qty(\bm{x}_s, \qty{y_s, \bm{b}_s})\sim\mathcal{D}_s} 
    \qty[\mathcal{L}_{\mathrm{det}}\qty(F\qty(\bm{x}_s + \bm{\delta^*}), \qty{y_s, \bm{b}_s})],
\end{align}
where $\bm{\delta^*}$ represents adversarial perturbation.
$\bm{\delta^*}$ is designed to cause the detector to make mistakes, and it is usually too small to be perceived by humans. 
Therefore, as shown in (\ref{eq:at}), the detector is dependent on the robust features that are aligned with human perception.
We shall now introduce several designs of perturbations $\bm{\delta^*}$ for YOLOv3 based on the FGSM~\cite{goodfellow2015explaining} and PGD~\cite{madry2018towards}. 
We shall then describe $\bm{\delta^*}$ used in our experiments.

\subsubsection{FGSM.}
The FGSM creates an adversarial perturbation in a single gradient step.
A straightforward approach for generating an adversarial perturbation involves using the gradient of $\mathcal{L}_{\mathrm{det}}$ as follows:
\begin{align}
    \label{eq:delta_ldet}
    \tilde{\bm{\delta}}_{\mathrm{det}} &= \sign \qty(\nabla_{\bm{\delta}_{0}} \mathcal{L}_{\mathrm{det}}\qty(F\qty(\bm{x}_s + \bm{\delta}_{0}), \qty{y_s, \bm{b}_s})),\\
    \label{eq:final_perturbation}
    \bm{\delta}_{\mathrm{det}} &= \mathcal{P}\qty[\bm{\delta}_{0} + \epsilon \cdot  \tilde{\bm{\delta}}_{\mathrm{det}}],
\end{align}
where $\mathcal{P}$ denotes the projection onto the $L_{\infty}$-norm $\epsilon$-ball $\qty{\bm{\delta}\mid\norm{\bm{\delta}}_\infty\leq\epsilon}$ for some $\epsilon > 0$, and $\bm{\delta}_{0}$ represents the initial value of the perturbation. 
As shown in (\ref{eq:delta_ldet}), $\tilde{\bm{\delta}}_{\textrm{det}}$ is calculated as a signed gradient of $\mathcal{L}_{\mathrm{det}}$ with respect to $\bm{\delta}_{0}$.
The adversarial perturbation $\bm{\delta}_{\textrm{det}}$ is then obtained using (\ref{eq:final_perturbation}).

Alternatively, one can generate adversarial perturbations $\bm{\delta}_{\mathrm{cls}}$, $\bm{\delta}_{\mathrm{loc}}$, and $\bm{\delta}_{\mathrm{obj}}$ based on the three task losses presented in (\ref{eq:detection_loss}) in a similar manner. 
First, $\tilde{\bm{\delta}}_{\mathrm{cls}}$, $\tilde{\bm{\delta}}_{\mathrm{loc}}$, and $\tilde{\bm{\delta}}_{\mathrm{obj}}$ are generated as follows:
\begin{align}
    \tilde{\bm{\delta}}_{\mathrm{cls}} &= \sign \qty(\nabla_{\bm{\delta}_{0}} \mathcal{L}_{\mathrm{cls}}\qty(F\qty(\bm{x}_s + \bm{\delta}_{0}), y_s)), \\
    \tilde{\bm{\delta}}_{\mathrm{loc}} &= \sign \qty(\nabla_{\bm{\delta}_{0}} \mathcal{L}_{\mathrm{loc}}\qty(F\qty(\bm{x}_s + \bm{\delta}_{0}), \bm{b}_s)), \\
    \tilde{\bm{\delta}}_{\mathrm{obj}} &= \sign \qty(\nabla_{\bm{\delta}_{0}} \mathcal{L}_{\mathrm{obj}}\qty(F\qty(\bm{x}_s + \bm{\delta}_{0}))). 
\end{align}
The final perturbations are then obtained as shown in (\ref{eq:final_perturbation}).
From the perspective of multi-task learning in object detection, \cite{zhang2019towards} showed that the direct use of $\mathcal{L}_{\mathrm{det}}$, as shown in (\ref{eq:delta_ldet}), results in gradient misalignment between tasks, thereby causing decreased robustness against adversarial examples.
To avoid this problem, they proposed an adversarial training method, which selects a single task perturbation that maximizes $\mathcal{L}_{\mathrm{det}}$.
Hereinafter, we denote this perturbation as $\bm{\delta}_{\mathrm{mtl}}$, which is generated for YOLOv3 as follows:
\begin{align}
\label{eq:delta_mtl}
    \bm{\delta}_{\mathrm{mtl}} &= 
    \argmax_{\bm{\delta} \in \qty{\bm{\delta}_{\mathrm{cls}}, \bm{\delta}_{\mathrm{loc}}, \bm{\delta}_{\mathrm{obj}}}} 
    \mathcal{L}_{\mathrm{det}}\qty(F\qty(\bm{x}_s + \bm{\delta}), \qty{y_s, \bm{b}_s}).
\end{align}

Generally, $\bm{\delta}_{0}$ is set to zero for the FGSM, which is referred to as the zero-initialized FGSM in this study. 
However, a recent study showed that initializing $\bm{\delta}_{0}$ using a random value uniformly sampled from $\qty[-\epsilon, \epsilon]$ results in enhanced robustness against adversarial examples~\cite{wong2020fast}.
We refer to this as the random-initialized FGSM.

\subsubsection{PGD.}
PGD generates stronger perturbations than those generated using the FGSM by iterating the gradient steps.
Adversarial training using PGD is known to be effective in enhancing adversarial robustness. 
However, this approach is computationally expensive.
With a step size parameter $\alpha$, the generation of adversarial perturbation using PGD can be expressed as follows:
\begin{align}
    \bm{\delta}^{\qty(t+1)} = \mathcal{P}\qty[\bm{\delta}^{\qty(t)} + \alpha \cdot \tilde{\bm{\delta}}^{\qty(t)}].
\end{align}
Here, $\tilde{\bm{\delta}}^{\qty(t)}$ can be computed using arbitrary losses in object detection, as described for the FGSM.

\subsubsection{Default perturbation $\bm{\delta^*}$ in our experiments.}
We employ $\bm{\delta}_{\mathrm{det}}$ generated using the zero-initialized FGSM as the default $\bm{\delta^*}$ in our main experiments, because this is the simplest strategy for adversarial training. 
In Section~\ref{subsec:analysis}, comparisons among the zero-initialized FGSM, random-initialized FGSM, and PGD are conducted, as well as a comparison of the losses used to generate perturbations.
Although PGD, the random-initialized FGSM, and the use of $\bm{\delta}_{\mathrm{mtl}}$ are known to enhance robustness to adversarial examples \cite{madry2018towards,wong2020fast,zhang2019towards}, we establish that the simplest adversarial training method, i.e., the zero-initialized FGSM with $\bm{\delta}_{\mathrm{det}}$, is sufficient in terms of performance with respect to domain adaptation.

\subsection{Robust and target-aligned feature learning}
\label{subsec:at_afl}
Through adversarial training in the source domain, as described above, the model is expected to learn robust features that are also informative to the target domain.
However, since the model is not trained in the target domain, the robust features acquired through the model are discrepant from the robust features in the target domain.
Therefore, we aim to enhance domain adaptation performance by aligning the robust features to the target domain.

For this purpose, we incorporate adversarial feature learning, which is a typical approach for implementing domain adaptation.
Specifically, we employ a local feature alignment approach that matches features, such as texture and color, between the source and target domains~\cite{saito2019strong-weak}.
In this study, the detector $F$ is decomposed as follows: $F_2 \circ F_1$, where $F_1$ represents the first dozens of the network layers, and $F_2$ represents the rest of the layers in the network.
The output of $F_1$ is the input of the domain discriminator $D$ across the gradient reversal layer~\cite{ganin2016domain-adversarial}.
The feature extractor $F_1$ outputs a feature map of width $W$ and height $H$, and the domain discriminator $D$ outputs a domain prediction map whose width and height are the same as those of the input from $F_1$.
In our setting, the domain discriminator $D$ aims to ensure that the domain predictions for the source images are equal to zero and that those for the target images are equal to one.
In contrast, the feature extractor $F_1$ is trained in a manner that ensures the domain predictions are opposite to those $D$ aims for.
Owing to the gradient reversal layer, the losses of adversarial feature learning can be summarized as follows:
\begin{align}
\label{eq:afl_s}
    \mathcal{L}_{\mathrm{afl}_s}\qty(D\qty(F_{1}\qty(\bm{x}_s + \bm{\delta^*})))
    &= \frac{1}{WH}\sum_{w,h} D\qty(F_{1}\qty(\bm{x}_s + \bm{\delta^*}))_{wh}^{2},\\
    \mathcal{L}_{\mathrm{afl}_t}\qty(D\qty(F_{1}\qty(\bm{x}_t))) 
    &= \frac{1}{WH}\sum_{w,h} \qty(1 - D\qty(F_{1}\qty(\bm{x}_t))_{wh})^{2},
\end{align}
where $D\qty(\cdot)_{wh}$ denotes the $\qty(w, h)$-th entry of the outputs of $D$.
Note that we add an adversarial perturbation $\bm{\delta^*}$ to the source image $\bm{x}_s$, as shown in (\ref{eq:afl_s}).

Combined with the objective of adversarial training in the source domain, (\ref{eq:at}), the overall objective is expressed as follows:
\begin{align}
\label{eq:at_afl}
    \max_{F_1}&\min_{F,D} \mathbb{E}_{\substack{\qty(\bm{x}_s, \qty{y_s, \bm{b}_s})\sim\mathcal{D}_s\\ \bm{x}_t\sim\mathcal{D}_t}}
    [\mathcal{L}_{\mathrm{det}}\qty(F\qty(\bm{x}_s + \bm{\delta^*}), \qty{y_s, \bm{b}_s}) \notag\\
    &+ \lambda \qty(\mathcal{L}_{\mathrm{afl}_s}\qty(D\qty(F_{1}\qty(\bm{x}_s + \bm{\delta^*}))) + \mathcal{L}_{\mathrm{afl}_t}\qty(D\qty(F_{1}\qty(\bm{x}_t))))].
\end{align}
where $\lambda$ represents the weight required to ensure the balance between adversarial training and adversarial feature learning.
The signs of the gradients back-propagated from $D$ to $F_1$ are reversed through the gradient reversal layer.

\section{Experiments}
In this section, we demonstrate the effectiveness of our approach through domain adaptation experiments conducted on benchmark datasets.
In addition, we compare various methods and parameters for generating adversarial perturbations in adversarial training, and we observe the effect of the choice on the domain adaptation performance.

\begin{figure}[t]
\centering
\includegraphics[width=0.45\textwidth]{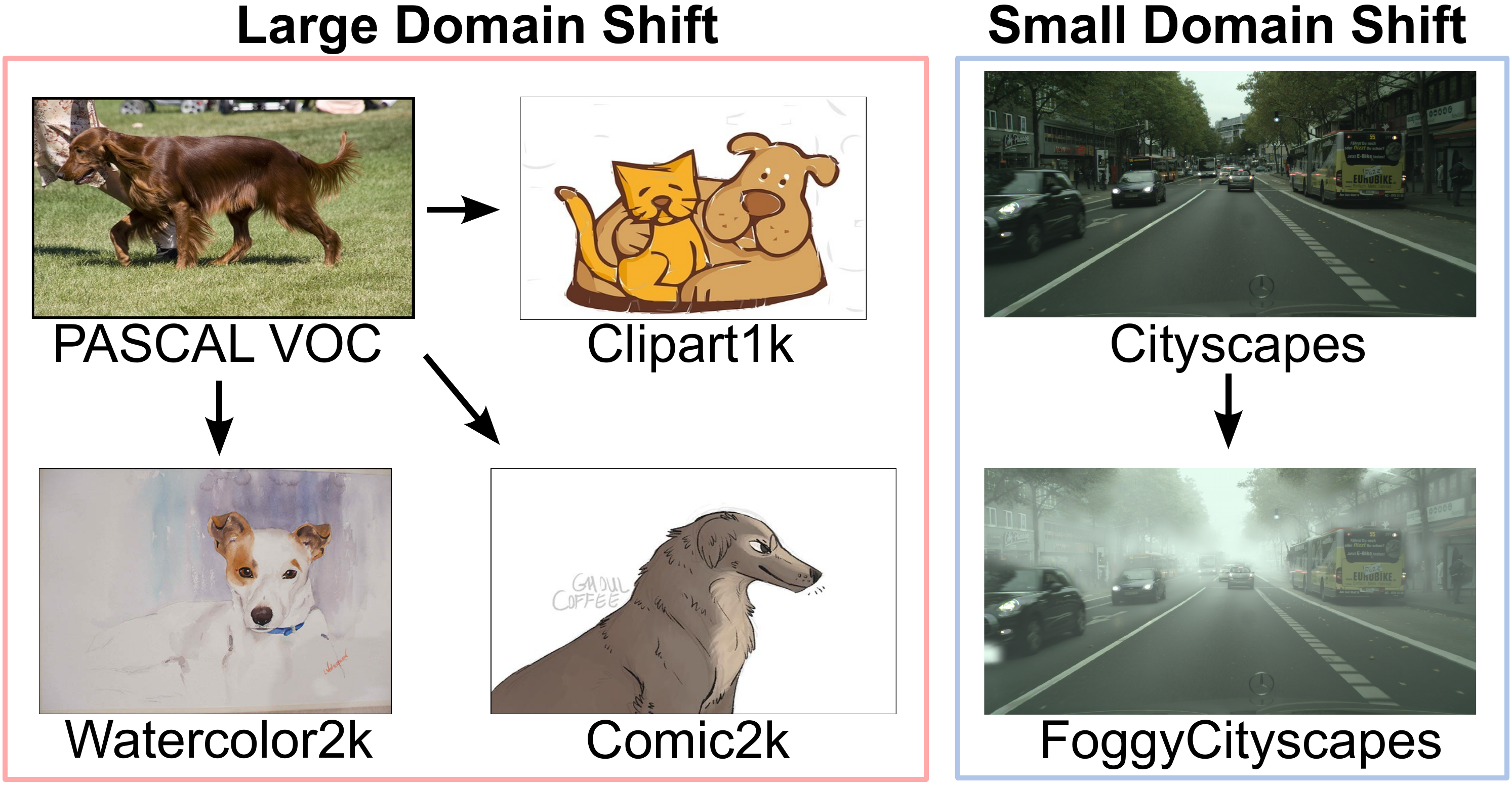}
\caption{
Examples of the datasets used in the experiments.
}
\label{fig:fujii3}
\end{figure}

\begin{table*}[t]
\caption{
Results of adaptation from PASCAL VOC to Clipart1k. 
AP (\%) is reported on the Clipart1k test set. 
ST, AT, and AFL indicate standard training, adversarial training, and adversarial feature learning, respectively.
}
\centering
\fontsize{8}{10}\selectfont
\setlength{\tabcolsep}{2pt}
\begin{tabular}{c|cccccccccccccccccccc|c}
    \hline
    Method & Aero & Bike & Bird & Boat & Bottle & Bus & Car & Cat & Chair & Cow & Table & Dog & Horse & M-bike & Person & Plant & Sheep & Sofa & Train & TV & mAP\\ \hline
    ST & 29.9 & 54.7 & 22.9 & 32.7 & 50.1 & 43.6 & 38.7 & 7.4 & 59.8 & 35.6 & 32.1 & 11.9 & 38.9 & 37.1 & 51.6 & \textbf{60.0} & 9.8 & 49.4 & 35.0 & 50.9 & 37.6\\
    ST + AFL & \textbf{33.9} & 65.4 & 26.6 & \textbf{43.7} & 58.2 & 45.5 & 48.6 & 7.3 & 60.7 & 51.8 & 40.6 & 15.1 & 36.6 & 51.4 & 59.8 & 58.8 & 19.4 & 51.2 & 44.2 & 55.5 & 43.7\\
    AT (ours) & 27.3 & 63.8 & 27.3 & 37.4 & 66.2 & 54.0 & 41.3 & \textbf{12.0} & 61.8 & 56.7 & 40.3 & 16.1 & 46.7 & 53.1 & 60.2 & 51.4 & \textbf{28.9} & \textbf{64.7} & 50.0 & 47.7 & 45.3\\
    AT + AFL (ours) & 28.9 & \textbf{67.0} & \textbf{28.9} & 40.6 & \textbf{67.6} & \textbf{69.0} & \textbf{57.1} & 10.6 & \textbf{64.2} & \textbf{62.6} & \textbf{42.1} & \textbf{18.3} & \textbf{51.0} & \textbf{55.3} & \textbf{65.6} & 53.5 & 28.5 & 64.4 & \textbf{54.3} & \textbf{58.7} & \textbf{49.4 }\\
    \hline
\end{tabular}
\label{tab:clipart}
\end{table*}

\begin{table}[t]
\caption{
Results of adaptation from PASCAL VOC to Watercolor2k. 
AP (\%) is reported on the Watercolor2k test set.
}
\centering
\fontsize{8}{10}\selectfont
\setlength{\tabcolsep}{3pt}
\begin{tabular}{c|cccccc|c}
    \hline
    Method & Bike & Bird & Car & Cat & Dog & Person & mAP \\ \hline
    ST & 90.9 & 49.5 & 51.9 & 27.0 & 16.5 & 59.4 & 49.2 \\
    ST + AFL & 78.9 & 44.0 & 48.9 & 24.8 & 14.1 & 55.1 & 44.3\\
    AT (ours) & \textbf{97.9} & 53.5 & 54.8 & \textbf{36.8} & 25.0 & 63.7 & 55.3\\
    AT + AFL (ours) & 96.1 & \textbf{54.0} & \textbf{56.4} & 35.4 & \textbf{27.2} & \textbf{64.6} & \textbf{55.6}\\
    \hline
\end{tabular}
\label{tab:watercolor}
\end{table}

\begin{table}[t]
\caption{
Results of adaptation from PASCAL VOC to Comic2k. 
AP (\%) is reported on the Comic2k test set.
}
\centering
\fontsize{8}{10}\selectfont
\setlength{\tabcolsep}{3pt}
\begin{tabular}{c|cccccc|c}
    \hline
    Method & Bike & Bird & Car & Cat & Dog & Person & mAP \\ \hline
    ST & 50.5 & 12.6 & 33.3 & \textbf{10.6} & 9.8 & 43.6 & 26.7 \\
    ST + AFL & 55.3 & \textbf{15.5} & 37.7 & 10.3 & 18.6 & 51.8 & 31.5 \\
    AT (ours) & 56.8 & 12.8 & 31.7 & 9.0 & 19.7 & 47.2 & 29.5 \\
    AT + AFL (ours) & \textbf{57.1} & 15.4 & \textbf{41.3} & 10.4 & \textbf{23.2} & \textbf{53.0} & \textbf{33.4} \\
    \hline
\end{tabular}
\label{tab:comic}
\end{table}

\subsection{Datasets}
For large domain shifts, we use PASCAL VOC~\cite{everingham2010pascal} as the source dataset and Clipart1k, Watercolor2k, and Comic2k~\cite{inoue2018cross} as the target datasets.
PASCAL VOC is a dataset comprising real-world images with 20 object classes. 
The training sets (VOC2007-trainval and VOC2012-trainval) comprise 16,551 images, and the test set (VOC2007-test) comprises 4,952 images.
Clipart1k is a dataset comprising graphical images and has the same object classes as PASCAL VOC. 
The training set comprises 500 images, and the test set comprises 500 images.
Watercolor2k and Comic2k are datasets comprising watercolor and comic images, respectively.
Both datasets have six object classes, which are defined in PASCAL VOC, and they comprise 1,000 training and 1,000 test images.
The appearances of objects significantly differ between the real images in the PASCAL VOC dataset and the artistic images in the Clipart1k, Watercolor2k, and Comic2k datasets.

For small domain shifts, we use Cityscapes~\cite{cordts2016cityscapes} as the source dataset and FoggyCityscapes~\cite{sakaridis2018semantic} as the target dataset.
Cityscapes is a dataset comprising urban street scenes with eight object classes.
The training set comprises 2,975 images, and the test set comprises 500 images.
FoggyCityscapes is a dataset rendered from Cityscapes with fog simulation; it comprises the same number of images as the Cityscapes dataset.
The weather conditions are different in the two datasets, but the appearances of the objects are similar.
Examples of the datasets are shown in Fig.~\ref{fig:fujii3}.

\subsection{Implementation details}\label{subsec:implementation details}
In this study, we use YOLOv3~\cite{redmon2018yolov3}, which is a well-known object detector.
The network with the first 26 convolutional layers of Darknet-53 in YOLOv3 is used as the feature extractor $F_1$, which is introduced in Section~\ref{subsec:at_afl}, and the rest of the network is used as $F_2$.
The domain discriminator $D$ is designed following the original local domain classifier~\cite{saito2019strong-weak}.
The training images are applied using Mosaic data augmentation~\cite{bochkovskiy2020yolov4} and resized to $416\times416$ pixels.
In all the experiments, a model pre-trained using the MSCOCO~\cite{lin2014microsoft} dataset is used as the initial weight.
We train the models for 50 epochs on the size of the source dataset.
The optimizer is a stochastic gradient descent with a momentum of 0.937 and a weight decay of $5.0 \times 10^{-4}$.
The learning rate decreases from $1.0 \times 10^{-2}$ to $2.0 \times 10^{-3}$ through the cosine annealing schedule, and linear warmup is used for the first three epochs.

During evaluation, the test images are resized, so that the longer side is 416.
We evaluate the average precision (AP) and mean AP (mAP) on the test data using an IoU threshold of 0.5.
The reported results are the average of over three runs of similar training procedures.
All the experiments are implemented using the PyTorch framework installed on the Ubuntu operating system running on a computer with an NVIDIA TITAN RTX GPU.

\subsubsection{Standard training.}
Only the source dataset is used, and the batch size is set to 16.
We add a zero tensor to the source image instead of $\bm{\delta^*}$, which is presented in (\ref{eq:at}).

\subsubsection{Adversarial training.}
Only the source dataset is used, and the batch size is similar to that for standard training.
By default, $\bm{\delta}_{\mathrm{det}}$ generated using the zero-initialized FGSM with $\epsilon=1/255$ is used as $\bm{\delta^*}$.
For detailed analysis, the method and parameters for generating $\bm{\delta^*}$ are modified as required.

\subsubsection{Adversarial feature learning.}
When adversarial feature learning is combined, both the source and target datasets are used.
The batch size is set to 32: 16 from the source dataset and 16 from the target dataset.
We set $\lambda=1.0$ in (\ref{eq:at_afl}).

\subsection{Results}
\label{subsec:result}

\subsubsection{Large domain shift.}
We first conduct experiments on adaptations in large domain shifts from real to artistic images.
Specifically, adaptations from PASCAL VOC to Clipart1k, Watercolor2k, and Comic2k are evaluated.

First, we list the results on the Clipart1k dataset in Table~\ref{tab:clipart}.
Adversarial training (AT) outperforms standard training (ST) by 7.7\% in terms of mAP.
In addition, AT outperforms ST combined with adversarial feature learning (ST+AFL) by 1.6\%, even though the target dataset is not used for AT.
AT combined with AFL (AT+AFL) outperforms the other methods for 14 classes in terms of AP and improves the mAP by 11.8\% compared to that of ST.
Next, we list the results on the Watercolor2k dataset in Table~\ref{tab:watercolor}.
AT and AT+AFL improve the mAP over that of ST by 6.1\% and 6.4\%, respectively.
AT+AFL outperforms the other methods for four classes in terms of AP, although the improvement achieved through AT is limited compared to that on other datasets.
Finally, we list the results on the Comic2k dataset in Table~\ref{tab:comic}.
AT and AT+AFL improve the mAP over that of ST by 2.8\% and 6.7\%, respectively.
AT+AFL outperforms the other methods for four classes in terms of AP.

In summary, adversarially trained models outperform standard–trained models, and further improvements in their performance can be achieved by incorporating AFL.
Specifically, the finding that AT using only the source dataset results in improved performance in the target domain is interesting because general domain adaptation methods utilize images in the target domain.
The reason behind these results can be explained as follows. 
In the adaptation of real to artistic images, the non-robust features acquired through ST in the source domain are not informative in the target domain owing to the large domain shift. 
As a result, ST degrades performance in the target domain.
Contrarily, the robust features acquired through AT are informative in the target domain, and thus, they can maintain performance in the target domain.
Combined with AFL, the robust features are aligned with the target domain, thereby resulting in further performance improvement.

\begin{table*}[t]
\caption{
Results of adaptation from Cityscapes to FoggyCityscapes. 
AP (\%) is reported on the FoggyCityscapes test set.
}
\centering
\fontsize{8}{10}\selectfont
\setlength{\tabcolsep}{3pt}
\begin{tabular}{c|cccccccc|c}
    \hline
    Method & Bike & Bus & Car & M-bike & Person & Rider & Train & Truck & mAP \\ \hline
    ST & 26.6 & 31.4 & 44.1 & 18.3 & 32.5 & 35.7 & 8.7 & 20.3 & 27.2 \\
    ST + AFL & \textbf{32.4} & \textbf{35.8} & \textbf{48.0} & \textbf{20.9} & \textbf{35.7} & \textbf{38.4} & \textbf{13.3} & \textbf{24.3} & \textbf{31.1} \\
    AT (ours) & 10.7 & 3.3 & 21.8 & 2.8 & 15.3 & 12.6 & 4.5 & 4.6 & 9.5 \\
    AT + AFL (ours) & 14.9 & 10.5 & 28.3 & 6.4 & 19.9 & 19.2 & 1.5 & 7.3 & 13.5 \\
    \hline
\end{tabular}
\label{tab:foggy}
\end{table*}

\subsubsection{Small domain shift.}
We also evaluate the effectiveness of our proposed method on a small domain shift.
Specifically, an adaptation between different weather conditions, from Cityscapes to FoggyCityscapes, is performed.
The results are listed in Table~\ref{tab:foggy}.
Contrary to the results for large domain shifts, AT and AT+AFL decrease the mAP by 17.7\% and 13.7\% compared to ST, respectively.
In this adaptation scenario, the ST+AFL approach demonstrates the best performance in all the classes in the target domain.

For adaptation between similar domains, the non-robust features acquired through ST are also informative in the target domain.
Therefore, the detection performance of standard-trained models in the target domain is highly dependent on non-robust features.
In contrast, AT makes the detector rely on robust features instead of non-robust features.
Because robust features are less informative than non-robust features, AT is known to result in a reduction in accuracy in the source domain \cite{tsipras2019robustness}.
Correspondingly, for small domain shifts, the application of AT results in decreased performance in the target domain.

\subsection{Analysis}
\label{subsec:analysis}

\begin{figure*}[t]
\centering
\includegraphics[width=0.99\textwidth]{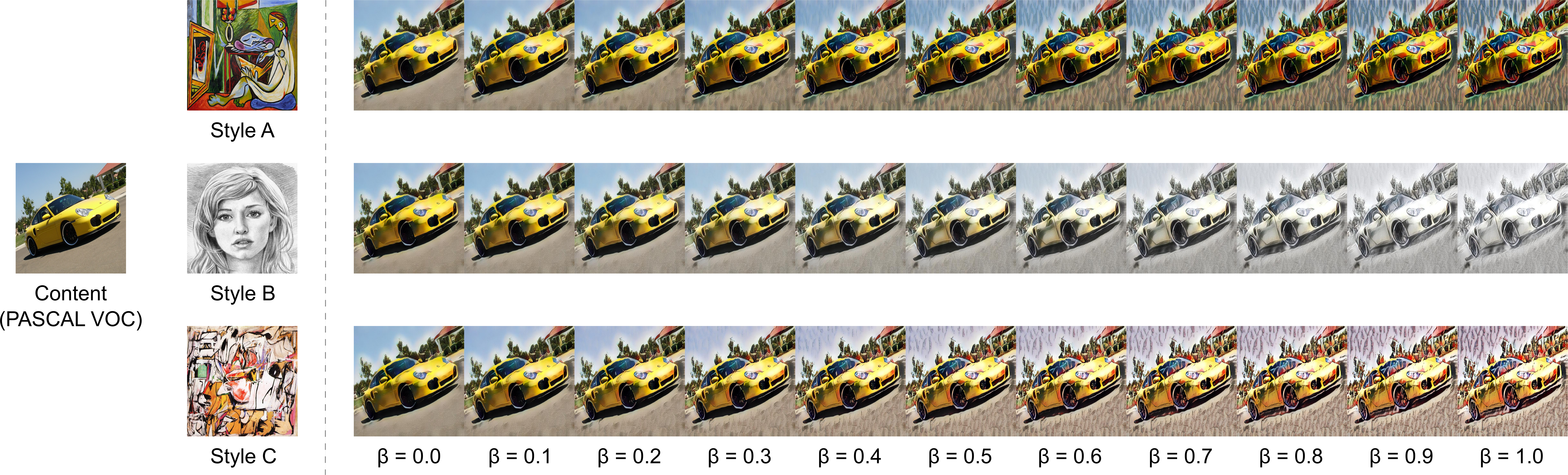}
\caption{
Examples of style transfer via AdaIN for the PASCAL VOC test set.
The test set is stylized into three style images by varying the content--style trade-off $\beta$.
}
\label{fig:fujii4}
\end{figure*}

\begin{figure*}[t]
\centering
\includegraphics[width=0.85\textwidth]{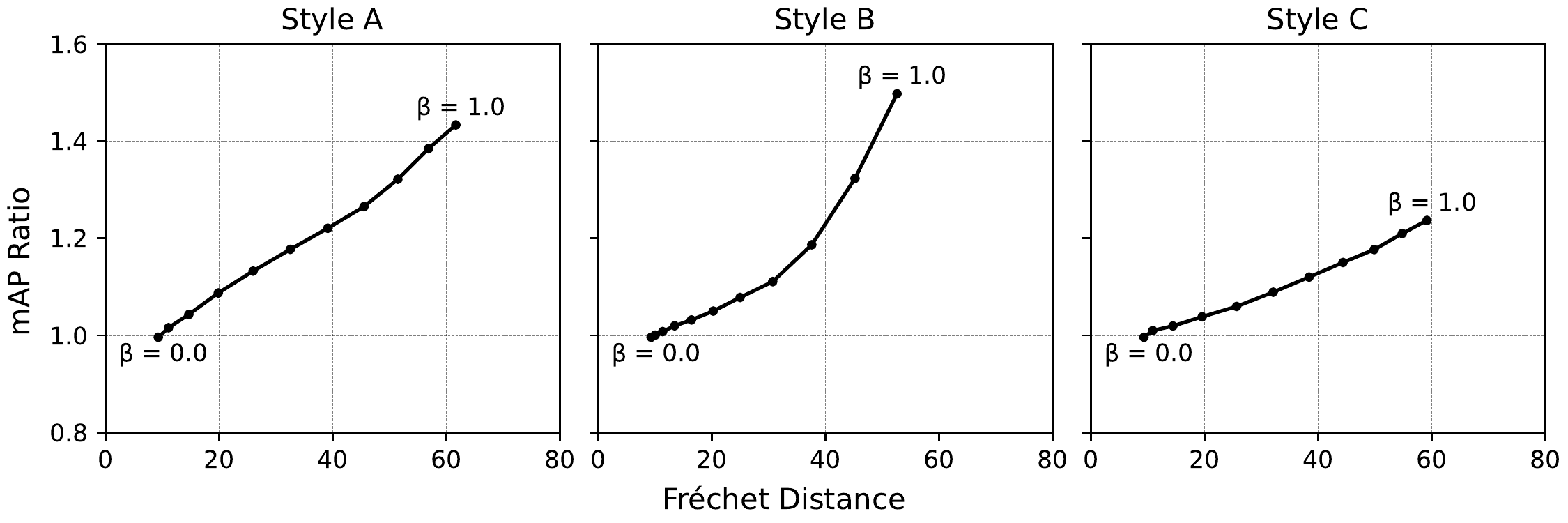}
\caption{
Fr\'echet distance (FD) between the PASCAL VOC training set and the stylized PASCAL VOC test set, and the ratio of the mAP of AT to mAP of ST on the stylized test set.
The larger the FD, the better the detection performance of AT on the stylized test sets.
}
\label{fig:fujii5}
\end{figure*}

\begin{table}[t]
\caption{
FD between the training set of the source domain and the test set of the target domain, and ratio of the mAP of AT to that of ST in the target domain.
}
\centering
\fontsize{8}{10}\selectfont
\setlength{\tabcolsep}{3pt}
\begin{tabular}{cc|c|c}
    \hline
    Source & Target & FD & mAP ratio\\ \hline
    Pascal VOC & Clipart1k & 44.2 & 1.20\\
    Pascal VOC & Watercolor2k & 42.1 & 1.12\\
    Pascal VOC & Comic2k & 59.0 & 1.10\\
    Cityscapes & FoggyCityscapes & 9.6 & 0.35\\
    \hline
\end{tabular}
\label{tab:fd}
\end{table}

\begin{figure*}[t]
\centering
\includegraphics[width=0.9\textwidth]{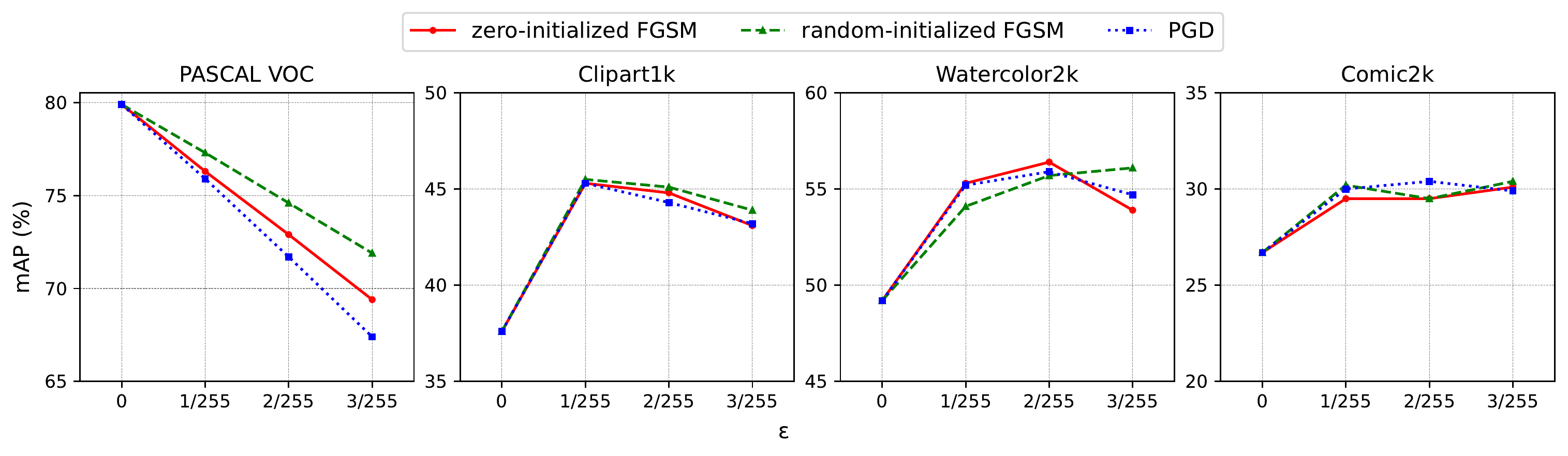}
\caption{
Results of AT using the zero-initialized FGSM, random-initialized FGSM, and PGD on the PASCAL VOC dataset with different values of $\epsilon$.
We report the mAP values on the PASCAL VOC, Clipart1k, Watercolor2k, and Comic2k test sets.
}
\label{fig:fujii6}
\end{figure*}

\subsubsection{Quantifying the domain shift magnitude.}
AT in the source domain improves performance for a large shifted target domain, but degrades performance for a small shifted target domain.
Therefore, quantifying the magnitude of the domain shift is necessary to determine whether our approach should be applied.
Fr\'echet inception distance (FID)~\cite{heusel2017gans}, which computes the Fr\'echet distance (FD) of two distributions using the feature space of the Inception-v3 model~\cite{szegedy2016rethinking}, is known as a measure to evaluate the difference between image sets.
We compute FD using the feature space of the YOLOv3 model instead of the Inception-v3 model, because it allows feature extraction along the object detection task.
In the experiments, we use feature maps extracted from the YOLOv3 backbone network, which is pre-trained using the MS COCO dataset.

First, we perform experiments on the PASCAL VOC dataset to verify the effectiveness of the FD to measure the domain shift.
To control the domain shift magnitude, we apply style transfer to the PASCAL VOC test set via adaptive instance normalization (AdaIN)~\cite{huang2017arbitrary}.
The content--style trade-off $\beta$, which manipulates the balance between content and style images, is varied from 0.0 to 1.0 in increments of 0.1.
Three style images used in our experiments and examples of the stylized images are shown in Fig.~\ref{fig:fujii4}.
The FD is computed between the original PASCAL VOC training set and each stylized PASCAL VOC test set.
We then conduct ST and AT on the training set and evaluate how much AT improves the mAP for each stylized test set compared to ST.
The results are shown in Fig.~\ref{fig:fujii5}.
As $\beta$ is increased, the FD between the training set and the stylized test set becomes larger.
Correspondingly, the ratio of the mAP of AT to that of ST also increases.
These results suggest that FD can be used to quantify the magnitude of the domain shift and help predict the effect of AT on domain adaptation.

Next, we compute the FD between the training set of the source domain and the test set of the target domain used in our main experiments.
The results are listed in Table~\ref{tab:fd}.
The FDs are larger for PASCAL VOC to Clipart1k, Watercolo2k, and Comic2k, where AT improves the detection performance in the target domain, and smaller for Cityscapes to FoggyCityscapes, where AT degrades the performance.
Thus, by measuring the magnitude of the domain shift in terms of FD, it is possible to determine whether adversarial training in the source domain should be conducted.

\subsubsection{Methods and parameters for adversarial training.}
AT using the random-initialized FGSM or PGD is known to make the model significantly robust to adversarial examples compared to AT using the zero-initialized FGSM~\cite{madry2018towards,wong2020fast}.
In addition, the value of $\epsilon$ is a crucial factor in AT.
In this study, we analyze the impact of the methods and parameters for AT on the performance of unsupervised domain adaptation.

We conduct AT using the zero-initialized FGSM, random-initialized FGSM, and PGD on the PASCAL VOC dataset using the gradient of $\mathcal{L}_{\mathrm{det}}$ and varying $\epsilon$.
When $\epsilon=0$, ST is performed instead of AT.
PGD is performed in 10 steps, and the step size is set to $\alpha=1.5 \epsilon / 10$.
Fig.~\ref{fig:fujii6} shows the mAP values for the source (PASCAL VOC) and target (Clipart1k, Watercolor2k, and Comic2k) test sets in each setting.
In the source domain, all the methods show a decrease in mAP as the value of $\epsilon$ increases. 
This is because AT prevents the model from acquiring predictive and non-robust features.
In the target domain, all the methods show an improvement in mAP compared to ST ($\epsilon=0$).
Interestingly, we establish that the three methods, known to differ in robustness against adversarial examples, are not substantially different in their performance in the target domain.
This result indicates the intriguing phenomenon that the domain adaptation performance of adversarially trained models does not depend on their robustness.
Considering the computational cost, the zero-initialized FGSM and random-initialized FGSM are better choices for domain adaptation.
On the other hand, the best value of $\epsilon$ depends on the target dataset and method; thus, $\epsilon$ must be adjusted according to the setting.

\begin{table}[t]
\caption{
Comparison of loss used for generating $\bm{\delta^*}$ using the zero-initialized FGSM. 
We report the mAP values (\%) for each target domain.
}
\centering
\fontsize{8}{10}\selectfont
\setlength{\tabcolsep}{3pt}
\begin{tabular}{c|ccc}
    \hline
     & \multicolumn{3}{c}{mAP on target datasets} \\
    $\bm{\delta^*}$ & Clipart1k & Watercolor2k & Comic2k\\ \hline
    $\bm{\delta}_{\mathrm{det}}$ & \textbf{45.3} & \textbf{55.3} & 29.5 \\
    $\bm{\delta}_{\mathrm{mtl}}$ & 45.1 & 55.0 & 29.1 \\
    $\bm{\delta}_{\mathrm{cls}}$ & 45.2 & 53.2 & \textbf{29.7} \\
    $\bm{\delta}_{\mathrm{loc}}$ & 44.4 & 54.0 & 28.3 \\
    $\bm{\delta}_{\mathrm{obj}}$ & 44.5 & 53.7 & 29.0 \\
    \hline
\end{tabular}
\label{tab:pertubation}
\end{table}

\subsubsection{Loss for generating adversarial perturbations.}
The total loss of object detection comprises several task losses, as shown in (\ref{eq:detection_loss}), for YOLOv3.
Therefore, determining the loss to be used to generate adversarial perturbations is a crucial factor.
To prevent gradient misalignment between tasks, the technique of selecting a single task loss that maximizes the total loss has also been proposed~\cite{zhang2019towards}, as shown in (\ref{eq:delta_mtl}).
We analyze the impact of these loss choices on the performance of domain adaptation.

We conduct AT using the zero-initialized FGSM on the PASCAL VOC dataset by varying the loss used to generate $\bm{\delta^*}$.
Table~\ref{tab:pertubation} shows the mAP values in the target domain for detectors trained using each adversarial perturbation.
AT using $\bm{\delta}_{\mathrm{det}}$ demonstrates the best performance for the Clipart1k and Watercolor2k datasets, and it is only 0.2\% lower than the best performance for the Comic2k dataset.
The use of $\bm{\delta}_{\mathrm{mtl}}$ is within only 0.6\% of the best performance on all datasets.
With $\bm{\delta}_{\mathrm{cls}}$, $\bm{\delta}_{\mathrm{loc}}$, and $\bm{\delta}_{\mathrm{obj}}$, which use a single task loss, the mAP values for the Watercolor2k dataset are much lower than the best performance by 1.3\% to 2.1\%.
These results suggest that domain adaptation performance is highly stable when all the task losses are considered during AT, as in the case of $\bm{\delta}_{\mathrm{det}}$ and $\bm{\delta}_{\mathrm{mtl}}$.
As mentioned in Section~\ref{subsec:at_in_source}, 
$\bm{\delta}_{\mathrm{mtl}}$ is known to be more robust than $\bm{\delta}_{\mathrm{det}}$ against adversarial examples because $\bm{\delta}_{\mathrm{det}}$ results in gradient misalignment between tasks, whereas $\bm{\delta}_{\mathrm{mtl}}$ does not.
However, $\bm{\delta}_{\mathrm{det}}$ shows a higher mAP value than $\bm{\delta}_{\mathrm{mtl}}$.
This indicates that the acquisition of robust features for domain adaptation must be considered separately from robustness against adversarial examples.

\section{Conclusion}
In this study, we explored the implementation of unsupervised domain adaptation in the field of object detection.
Further, we proposed a method based on adversarial training in the source domain. To the best of our knowledge, this is the first application of adversarial training in unsupervised domain adaptation.
The robust features acquired using adversarially trained detectors are informative in a largely shifted target domain, thereby resulting in improved detection performance.
In contrast, for small domain shifts where the non-robust features acquired through standard training are informative in both domains, adversarially trained detectors degrade performance in the target domain.
We also proposed a method for aligning the robust features with the target domain through adversarial feature learning and, using this approach, we demonstrated further improved performance for large domain shifts.

\clearpage

\bibliography{ref}

\end{document}